\documentclass{ecai} 
\usepackage{latexsym}
\usepackage{amssymb}
\usepackage{amsmath}
\usepackage{amsthm}
\usepackage{booktabs}
\usepackage{enumitem}
\usepackage{graphicx}
\usepackage{color}
\usepackage{xcolor}
\usepackage{multirow}
\usepackage[none]{hyphenat}
\usepackage{algorithm}
\usepackage{algorithmic}

\newtheorem{definition}{Definition}
\newtheorem{example}{Example}

\DeclareMathOperator*{\argmax}{arg\,max}
\DeclareMathOperator*{\argmin}{arg\,min}

\begin{document}

%%%%%%%%%%%%%%%%%%%%%%%%%%%%%%%%%%%%%%%%%%%%%%%%%%%%%%%%%%%%%%%%%%%%%%%%

\begin{frontmatter}

%%% Use this command to specify your submission number.
%%% In doubleblind mode, it will be printed on the first page.

\paperid{1788} 

%%% Use this command to specify the title of your paper.

\title{Pruning-Based Extraction of Descriptions from Probabilistic Circuits} %Deriving Comprehensible Theories from Probabilistic Circuits}

%%% Use this combinations of commands to specify all authors of your 
%%% paper. Use \fnms{} and \snm{} to indicate everyone's first names 
%%% and surname. This will help the publisher with indexing the 
%%% proceedings. Please use a reasonable approximation in case your 
%%% name does not neatly split into "first names" and "surname".
%%% Specifying your ORCID digital identifier is optional. 
%%% Use the \thanks{} command to indicate one or more corresponding 
%%% authors and their email address(es). If so desired, you can specify
%%% author contributions using the \footnote{} command.

\author[A]{\fnms{Sieben}~\snm{Bocklandt}\thanks{Corresponding Author. Email: sieben.bocklandt@kuleuven.be.}}
\author[B]{\fnms{Vincent}~\snm{Derkinderen}}
\author[C]{\fnms{Koen}~\snm{Vanderstraeten}} 
\author[D]{\fnms{Wouter}~\snm{Pijpops}} 
\author[E]{\fnms{Kurt}~\snm{Jaspers}} 
\author[F]{\fnms{Wannes}~\snm{Meert}} 

\address[A,B,F]{DTAI, KU Leuven}
\address[C,D,E]{Tunify}

%%% Use this environment to include an abstract of your paper.

\begin{abstract}
%Learning a concept from data is a general task with applications in various domains.
Concept learning is a general task with applications in various domains. 
As a motivating example we consider the application of music playlist generation, where a playlist is represented as a concept (e.g., `relaxing music') rather than as a fixed collection of songs. In this work we use a probabilistic circuit to learn a concept from positively labelled and unlabelled examples. While these circuits form an attractive tractable model for this task, it is challenging for a domain expert to inspect and analyse them, which impedes their use within certain applications.
We propose to resolve this by converting a learned probabilistic circuit into a logic-based discriminative model that covers the high density regions of the circuit. That is, those regions the circuit classifies as certainly being part of the learned concept. As part of this approach we present two contributions: PUTPUT, an algorithm to prune low density regions from a probabilistic circuit while considering both the F$_1$-score and a newly proposed description length that we call aggregated entropy.
Our experiments demonstrate the effectiveness of our approach in providing discriminative models, outperforming competitors on the music playlist generation task and similar datasets.

\end{abstract}
\end{frontmatter}

%%%%%%%%%%%%%%%%%%%%%%%%%%%%%%%%%%%%%%%%%%%%%%%%%%%%%%%%%%%%%%%%%%%%%%%%
\section{Introduction}

%https://www.zionmarketresearch.com/report/music-streaming-market?trk=article-ssr-frontend-pulse_little-text-block
%https://explodingtopics.com/blog/music-streaming-stats
%https://www.statista.com/outlook/dmo/digital-media/digital-music/music-streaming/worldwide
We motivate our work, which is more generally applicable, through an application of the music streaming industry.
This is a growing market, worth billions of USD~\cite{ecsa,unesco}. One important component of this industry are music playlists. From a user's perspective, they present a low barrier to discover- and automatically play a collection of related music. From a platform's perspective, the curation of playlists provides them with a mechanism to mediate markets and advance their own interests~\cite{Bonini2019,Prey20}. While many playlists are maintained by a human, a vast amount is curated automatically~\cite{Bonini2019}, forming an active line of research~\cite{AucouturierP02,BonninJ14,FlexerSGW08,LiuHT10,PichlZS16,SakuraiTOH22,Vall15}. 

A playlist is best represented as its concept (e.g., `relaxing music') rather than as a fixed collection songs, as it then automatically scales to consider newly released music.
%In terms of scaling and automatically considering newly released music, a playlist is best represented as its concept (e.g., `relaxing music') rather than as a fixed collection of songs. 
For instance, given some songs from a playlist, we could learn a discriminative model to represent the playlist's concept. The model can then automatically populate a playlist and consider additional songs as they are released.
%Given a set of songs from a playlist, one could learn a model to represent the playlist's concept. This allows us to automatically construct a playlist of related music and, in a similar manner, automatically decide whether newly released songs belong to the playlist as well. 
This learning process is typically a PU learning setting~\cite{PU}, where a few examples belonging to the concept are labelled positively and the rest are unlabelled.
%Naturally, this setting involves a dataset with many unlabelled examples. Furthermore, many if not all of the labelled instances will be positive, forming what is called a PU learning task~\cite{PU}. 

\begin{figure*}[t]
    \centering
    \includegraphics[width=\textwidth]{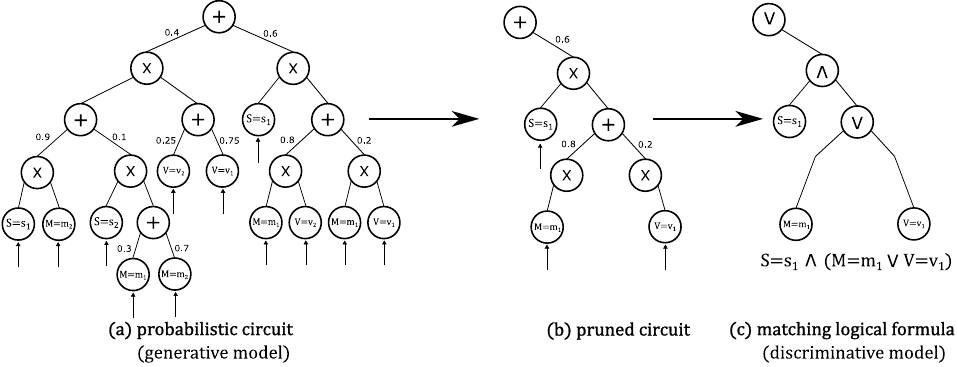}
    \caption{Overview of our approach. A (large) probabilistic circuit over variables \textit{Style}(S), \textit{Mood}(M) and \textit{Version}(V) is pruned to only contain the high density regions. This smaller circuit is transformed into a logic formula, acting as a discriminative model that is easier to inspect and verify by a domain expert.}
    \label{fig:intro-box}
\end{figure*}

% What are PCs, provide context
Probabilistic circuits (PC) have been introduced as a single unifying framework for tractable probabilistic models~\cite{circuits_ProbCircuits} such as arithmetic circuits~\cite{Darwiche03} and sum-product networks~\cite{PoonD11}. As a form of expressive deep generative models, they represent a joint probability distribution and are capable of tractably addressing a variety of tasks. Because of these capabilities, they present an interesting option to apply on our music playlist generation task. Figure~\ref{fig:intro-box}a shows an illustrative example of a learned probabilistic circuit in this context.

However, part due to their tractability, probabilistic circuits learned from data are harder to inspect and verify by a domain expert, which impedes their use within certain applications. For example, \textit{Tunify} provides a music streaming service specifically tailored towards businesses. As a consequence they must be able to assure their customers of the quality of each playlist. For instance, just one inappropriate song can ruin the atmosphere in a wellness centre or a funeral home. Therefore, ideally a domain expert can inspect the learned probabilistic circuit and verify that what it has learned to represent is indeed the intended concept and is safe for deployment.

% Solution.
We propose to extract a discriminative model from a probabilistic circuit, in a way that it is easier to inspect and understand by a domain expert. 
Importantly, we focus on the high density regions as the low density regions of a circuit represent examples (e.g., songs) that are less likely to be part of the intended concept.
For this reason, we developed a novel method that we call PUTPUT (Probabilistic circuit Understanding Through Pruning Underlying logical Theories). This method prunes the low density regions from the circuit (see Figure~\ref{fig:intro-box}b) while considering the aggregated entropy. This is a newly proposed description length that measures the ease with which a domain expert could inspect the extracted formula that acts as the discriminative model (see Figure~\ref{fig:intro-box}c). 

%Contributions
\textbf{The contributions} of this work are twofold: 
\begin{enumerate}
    \item it presents PUTPUT, a new pruning-based method to remove low density regions of a probabilistic circuit;
    \item it proposes a new description length called aggregated entropy to measure the ease with which a domain expert can inspect and understand a given logical formula; 
\end{enumerate}
The evaluation showcases the effectiveness of PUTPUT on the motivating real world use case of music playlist generation, where it outperforms competitors in terms of both F$_1$-score and aggregated entropy. Furthermore, evaluation on open source datasets shows that the method is more generally applicable outside of this use case.

\section{Theoretical background}
\label{Sec:theory}
We first provide a brief primer on logical formulas, probabilistic circuits and the used binary classification metrics.

\subsection{Logic formulas} 
A \emph{literal} $l$ is a Boolean variable $v$ or its negation $\neg v$. A \emph{logical formula} $\psi$ is inductively defined as a literal, the negation of logical formula $\neg \psi_1$, the conjunction (read `and') of two logical formulas $\psi_1 \land \psi_2$, or a disjunction (read `or') $\psi_1 \lor \psi_2$, with the expected semantics. A \emph{clause} $\mathcal{C}$ is a literal or disjunction of literals, such as $v_1 \lor \neg v_2$. A formula $\psi$ is said to be in \emph{conjunctive normal form} (CNF) iff it is a conjunction of clauses, such as $(v_1 \lor \neg v_2) \land (\neg v_1)$. A formula $\psi$ is said to be in \emph{disjunctive normal form} (DNF) iff it is a disjunction of conjunctions of literals, such as $(v_1 \land \neg v_2) \lor (\neg v_1)$.

We support categorical variables by admitting Boolean variables that represent equalities. That is, a Boolean variable $v$ could represent $style = jazz$, while another variable represents $style = rock$. We assume an implicit theory that forbids both variables to be true at the same time. For convenience, we may write $style = jazz$ rather than $v_{style=jazz}$.

An \emph{example} in our context is an assignment to each variable. 
% We represent an example as a set of variable-value pairs, e.g., $\{v_1 {=} \top,v_2 {=} \bot,...,v_n {=} \top\}$, where $\top$ means `true' and $\bot$ means `false'.

%Within probabilistic circuits and logical circuits we assume Boolean variables, so we encode categorical variables using a one-hot encoding resulting in a set of mutually exclusive Boolean variables. Given categorical variable $X$, the encoding in Boolean variables is given by $vals(X)=\{x_1,...,x_n\}$. 

The \emph{dual graph} $G_d(\psi)$ of a CNF formula $\psi$ is graph that connects clauses iff they share the same variable~\cite{SamerS10}. In our context we slightly change this definition to reason over categorical and binary variables. More formally, two clauses $\mathcal{C}$ and $\mathcal{C}'$ are joined by an edge if and only if there is a categorical or binary variable that is present in both clauses:
\begin{equation*}
    e(\mathcal{C},\mathcal{C}') \iff \exists X: X \in \mathcal{C} \text{ and } X \in \mathcal{C}' ,
    %\{X|X\in \mathcal{C}\} \cap \{X|X\in \mathcal{C}'\}\neq\emptyset ,
\end{equation*} using $X \in \mathcal{C}$ to denote categorical variables present in clause $\mathcal{C}$.

%we use this do denote categorical variables. the graph with the set of clauses in $\psi$ as vertex set \cite{SamerS10}. Two clauses $C$,$C'$ are joined by an edge if and only if there is a categorical variable that is present in both clauses: $e(C,C') \Leftrightarrow \{X|X\in C\} \cap \{X|X\in C'\}\neq\emptyset$ with $X\in C$ all categorical variables present in clause $C$ . 

\subsection{Probabilistic Circuits}\label{sec:bg:prob_circuits}

A probabilistic circuit (PC) $\mathcal{M}:=(\mathcal{G}, \theta)$ is a probabilistic model, representing a joint probability distribution P(\textbf{X}) over random variables \textbf{X} through a directed acyclic graph (DAG) $\mathcal{G}$ parameterised by $\theta$ \cite{circuits_ProbCircuits}. Each node in the DAG defines a computational unit, which is one of three types — an input, sum, or product node. Every leaf in $\mathcal{G}$ constitutes an input node, while every inner node is either a sum or a product node. An example of a probabilistic circuit is given in Figure~\ref{fig:intro-box}a.

% \begin{figure}[h!]
%     \centering
%     \includegraphics[width=.9\columnwidth]{figures/new_PC.pdf}
%     \caption{A probabilistic circuit with binary input variables Style (S), Version (V) and Mood (M).}
%     \label{fig:ProbCircuit}
% \end{figure}

In our setting, each input node of a PC represents a distribution over a categorical random variable $X$. Furthermore, we assume without loss of expressivity that the distribution has all probability mass over a single value $x \in X$. In other words, an input node labelled ${X=x}$ in Figure~\ref{fig:intro-box}a represents ${P(X=x|X=input)}$. 
%Each input node of a PC represents a univariate input distribution. 
Each product node represents a factorisation of incoming distributions over different random variables, and each sum node represents a mixture, i.e., a weighted sum over the distributions leading into it. The weights of the mixture are indicated on the edges in Figure~\ref{fig:intro-box}a. 
%In our work we do not consider continuous variables and assume the input distributions to be over discrete random variables $\mathbf{X}$. Furthermore, due to the mixture weights, we assume without loss of expressivity that each input node represents a categorical distribution with probability mass over a single value $x$. In other words, an input node labelled $X=x$ in Figure~\ref{fig:ProbCircuit} represents $P(X=x|X=input)$. 

\begin{figure}
    \centering
    \includegraphics[width=.9\columnwidth]{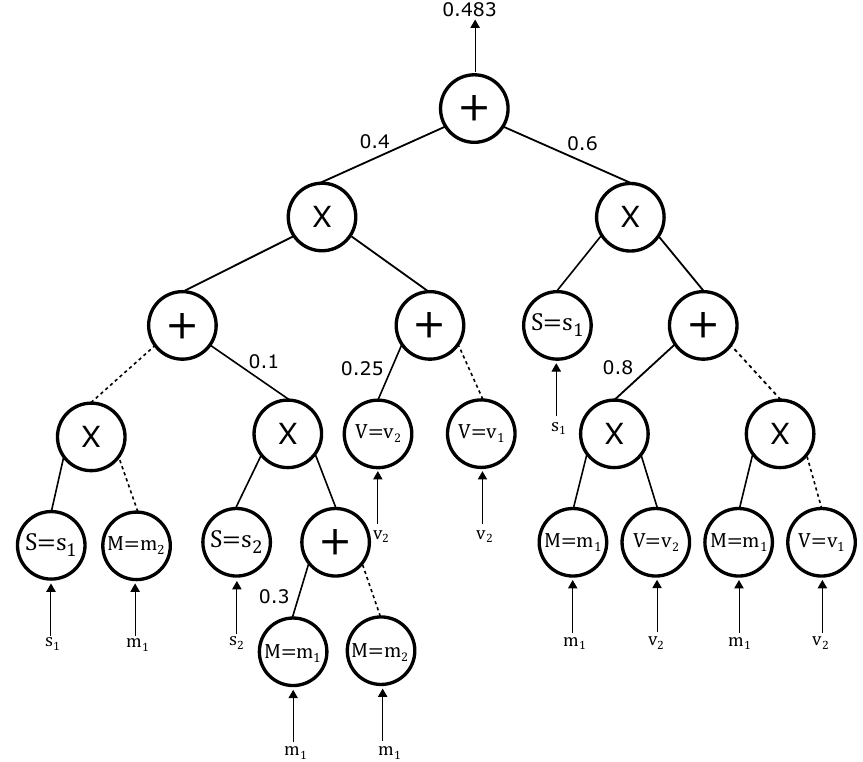}
    \caption{Generation of a probability with given input $\{m_1,v_2\}$. We marginalise over $S$, as no value is assigned. The dashed lines show the parts of the circuit where a zero probability is propagated.}
    \label{fig:evaluation}
\end{figure}

Figure~\ref{fig:evaluation} shows an evaluation of the PC to compute $P(Mood = m_1, Version = v_2)$, marginalising over $Style$. First, we set the input of the PC. For example, each input node associated to $M$ has input $m_1$, leading them to output $P(M=m_1|M=m_1) = 1$ or $P(M=m_2|M=m_1) = 0$.
We marginalise over $Style$, denoted as $S$, thus each input node $S{=}s$ receives input $s$ and outputs $1$.   
These outputs propagate through the PC, performing the sum- and product operations, resulting in the output probability at the root node. 

% \begin{equation}
%   p_n(\textbf{x}) = \begin{cases}
%   f_n(\textbf{x}) & \text{if n is an input node} \\
%   \Pi_{d\in in(n)}p_d(\textbf{x}) & \text{if n is a product node}\\ 
%   \sum_{d\in in(n)}\theta_{d|n}\cdot p_d(\textbf{x}) & \text{if n is a sum node}
%   \end{cases}
%   \end{equation}
% where $f_n(\textbf{x})$ is a univariate input distribution of a literal and $\theta_{d|n}$ denotes the nonnegative weight that corresponds to the edge $(d,n)$ in the DAG. The probability distribution of a probabilistic circuit is defined as the distribution represented by its root unit $p_{\mathcal{P}}(\textbf{x})$. The scope of a node is the set of input variables it depends on. A sum unit is smooth if the scopes of all the children are identical. A product unit is decomposable if the scopes of all the children are disjoint \cite{circuits_Pruning_growing}. 
% In this work, we limit the input distribution $f_n(\textbf{x})$ of a literal to be a numerical boolean function. 

\begin{figure}
    \centering
    \includegraphics[width=.9\columnwidth]{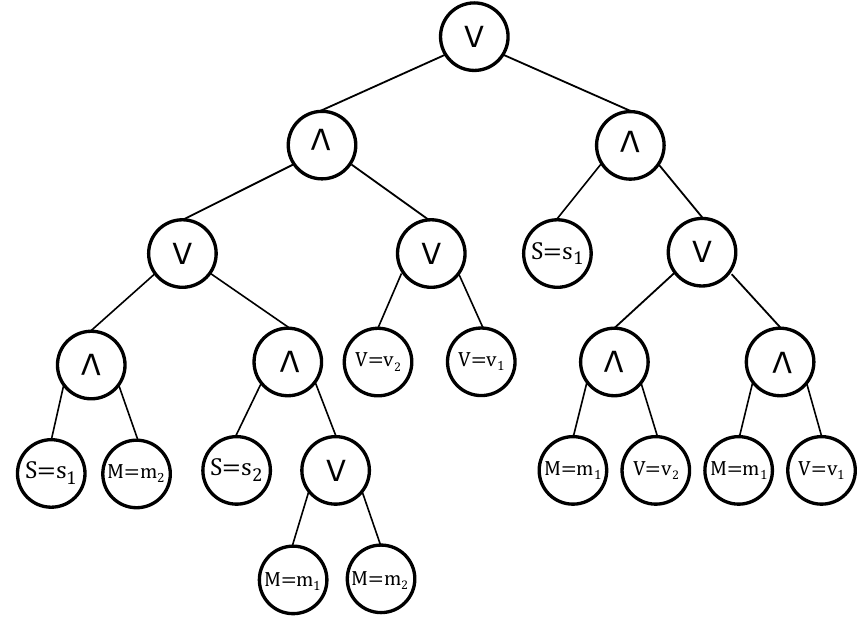}
    \caption{Logical formula derived from the probabilistic circuit in Figure~\ref{fig:intro-box}a.}
    \label{fig:logic} 
\end{figure}

A probabilistic circuit $\mathcal{M}$ can easily be converted to a logical formula $\psi$ that captures the input instances that have a nonzero probability within the circuit. Assuming only nonzero weights on the edges, $\psi$ can be extracted from $\mathcal{M}$ by replacing every product with $\land$, every sum node with $\lor$, and removing the weights of every edge. Additionally, the input nodes are converted into literals such as $S{=}s_1$. Figure~\ref{fig:logic} shows the logical formula $\psi$ extracted from Figure~\ref{fig:intro-box}a.

\subsection{Binary classification}

%\vincent{Move to experiment section, in more compact form.?}
Binary classification is the task of classifying an input example in one of two classes. To evaluate a binary classifier, functions such as F$_1$-score, precision and recall can be used. Precision is given by $\frac{\mathit{TP}}{\mathit{TP}+\mathit{FP}}$, where $\mathit{TP}$ are the examples that are correctly classified as positive and $\mathit{FP}$ the examples that are falsely classified as positive. It shows what fraction of the examples classified as positive are actually positive. Recall is given by $\frac{\mathit{TP}}{\mathit{TP}+\mathit{FN}}$ where $\mathit{FN}$ are the examples falsely classified as negative. Recall shows what fraction of the positive examples are correctly classified. These two can be combined as a single measure by taking the harmonic mean, resulting in the F$_1$-score $\frac{2\times\mathit{TP}}{2\times\mathit{TP}+\mathit{FP}+\mathit{FN}}$ that can be used to compare different classifiers. A higher F$_1$-score generally indicates that the classifier is more accurate when classifying.

%%%%%%%%%%%%%%%%%%%%%%%%%%%%%%%%%%%%%%%%%%%%%%%%%%%%%%%%%%%%%%%%%%%%%%%%

\section{A new description length for logic formulas in CNF}
\label{sec: comprehensibility}
The method that we will propose captures the high density regions of a probabilistic circuit as a logical formula that is easier to inspect and understand by a domain expert. To measure the latter, we introduce a new description length named \emph{aggregated entropy} that we describe after first motivating its introduction. 

\subsection{Motivation of aggregated entropy}
First, consider that nested logic formulas consisting of multiple levels of $\land$ and $\lor$ are more complex than flat forms such as CNF and DNF. Second, consider that DNFs are more complex to inspect once categorical variables are involved. We illustrate this using the following example from our application.
\begin{equation}
    \begin{split}
        (style=jazz &\land \mathit{feel}=happy) \lor \\
            (style=jazz &\land \mathit{feel}=exciting) \lor \\
            (style=rock &\land \mathit{feel}=happy) \lor \\
            (style=rock &\land \mathit{feel}=exciting)
    \end{split}
\end{equation}
In this case, a CNF representation is preferred.
\begin{equation}
    \begin{split}
        (style=jazz \lor style=rock) \land \\
        (\mathit{feel}=happy \lor \mathit{feel}=exciting)
    \end{split}
\end{equation}
% \vincent{A more refined suggestion might be to consider DNFs where a literal is instead allowed to be a set of values of a categorical variable. That might be a bit late, but good to think about a rebuttal if needed.}
An additional advantage of CNF formulas is that they can easily be extended to remove undesired examples $e$ (e.g., songs) as examples are conjunctions of attributes: $\psi \land \neg e$ which is equal to $\psi \land (\neg v_1 \lor \dots \lor \neg v_n)$.

Description lengths serve as a tool in various works, providing a quantitative measure of complexity or information content within data or models. DUCE~\cite{duce}, for instance, utilises the literal count within a formula to predict the impact of logical operators during the process of learning rules via constructive induction. Similarly, description length finds an application in information theory, as illustrated by the Huffman encoding which minimises the number of bits required to describe a sentence~\cite{Huffman}. Furthermore, Mistle~\cite{mistle} uses an encoding based on the universal code for integers alongside predicate invention to generate logical theories guided by the minimum description length principle. These description lengths all capture the length of the logical formula.

Naturally, a longer formula is more likely to be harder to inspect and understand. However, we argue these description lengths are suboptimal in the context of this work because they neither consider the complexity arising when variables are present in multiple clauses, nor do they consider the categorical variables. We therefore propose a new description length that we call aggregated entropy of a CNF formula.

\subsection{Aggregated entropy}
This new description length is based on information theory and keeps two principles in mind:
\begin{itemize}
    \item[--] A CNF formula is easier to understand when a variable is present in only a few clauses.
    \item[--] A CNF formula is easier to understand when a categorical variable allows either very few or many values. As an example, consider a formula that expresses that a music style is only allowed to be metal, or one that expresses everything except metal.
\end{itemize}

Aggregated entropy approximates this by quantifying the number of bits needed to represent a clause and its directly linked clauses as a proxy for how much a user needs to memorise when reading the model.

\begin{definition}[Entropy of a variable within a clause]
    The entropy $E_{\mathit{var}}(\mathcal{C},X)$ of a categorical variable $X$ within a clause $\mathcal{C}$ is defined as
\begin{equation}
    \begin{split}
        E_{\mathit{var}}(\mathcal{C},X) = 
            &-\frac{|\mathcal{C}(X)|}{|X|} 
                log_2\Big( \frac{|\mathcal{C}(X)}{|X|} \Big) \\
            &- \frac{|X| - |\mathcal{C}(X)|}{|X|} 
                log_2\Big( \frac{|X| - |\mathcal{C}(X)|}{|X|} \Big),
    \end{split}
\end{equation}
    with $\mathcal{C}(X)$ the set of Boolean variables that are both associated with categorical variable $X$ and present in clause $\mathcal{C}$, and $|X|$ the total number of possible values for $X$. In other words, $|\mathcal{C}(X)|/|X|$ is the fraction of possible values for $X$ that are mentioned within clause $\mathcal{C}$.
\end{definition}

%The aggregated entropy $DL_{cl}(\mathcal{C})$ of a clause $\mathcal{C}$ aggregates the entropy of the variables $X$ within it.
\begin{definition}[Aggregated entropy of a clause]
    The aggregated entropy $\mathit{DL}_{cl}(\mathcal{C})$ of a clause $\mathcal{C}$ is a description length that aggregates the entropy of the variables $X$ within it,
    \begin{equation}
        \mathit{DL}_{cl}(\mathcal{C}) = \sum_{X \in \mathcal{C}} E_{var}(\mathcal{C},X),
    \end{equation}
    where we use $X \in \mathcal{C}$ to consider the categorical variables $X$ that are present in clause $\mathcal{C}$. %We may overload this definition to $E_{cl}(\mathcal{C}, \mathbf{X})$ to indicate a restricted aggregation over the categorical variables that are in both $\mathcal{C}$ and in the given set $\mathbf{X}$.
\end{definition}

The ease with which a CNF formula $\psi$ is understood decreases when variables are present in multiple clauses. Therefore, while considering the aggregated entropy for each clause $\mathcal{C}$ within $\psi$, we also consider the aggregate of its neighbouring clauses, i.e., the clauses $\mathcal{C}'$ with whom $\mathcal{C}$ shares variables. 
\begin{definition}[Aggregated entropy of a CNF]
    The aggregated entropy $DL(\psi)$  of a CNF formula $\psi$ is defined as
    \begin{equation}
        \mathit{DL}(\psi) = \sum_{\mathcal{C} \in \psi} \left[ DL_{cl}(\mathcal{C}) +
\sum_{\substack{\mathcal{C}' \in \psi | e(\mathcal{C},\mathcal{C}') \in G_d(\psi)}} \mathit{DL}_{cl}(\mathcal{C}') \right],
    \end{equation}
    with $e(\mathcal{C},\mathcal{C}') \in G_d(\psi)$ denoting the clause neighbour relationships through the dual graph defined in Section~\ref{Sec:theory}. % and $Vars(\mathcal{C})$ representing the categorical variables in $\mathcal{C}$.
\end{definition}

%When many clauses in a theory include the same variable, the theory is more difficult to understand as the interactions between these clauses must be considered. This can be expressed by constructing a dual graph with each clause as a node \cite{DualGraph}. An edge in this graph is constructed between two nodes when they include the same variable, e.g. there is an edge $e(c_i,c_l)$ between clauses $c_i:x=1\vee z=3$ and $c_l:w=7\vee x=2$, as they both include variable $x$. 

%To read one clause, the expected effort is the incomprehensibility of that clause and all of the clauses it is linked with. The incomprehensibility of the entire theory is the sum of this value for all clauses. Formally, the incomprehensibility of a CNF theory $\psi=\bigwedge_{i=1..n}c_i$ with variables $V(\mathcal{C})$ and clausal graph $\mathcal{G}$ is then calculated as:   
%\begin{align}
%I(\psi) = \sum_{c \in \psi} \left[ \Upsilon(c, V(\psi)) +
%sum_{\substack{c' \in \psi | e(c,c') \in \mathcal{G})}}\Upsilon(c',V(\psi)) \right]    
%\end{align}

In the rest of this paper we minimise this metric $\mathit{DL}(\psi)$ in order to obtain more preferred CNF formulas $\psi$.
%maximizing comprehensibility is used to describe the minimisation of incomprehensibility.

\begin{example}
Given formula $\psi$ over categorical variables $\{A,B,X\}$ with $|A|= 5$, $|B| = 6$, and $|X| = 7$.
\begin{equation}
    \psi = 
    \underbrace{(a_1 \vee a_2 \vee a_3)}_{\mathcal{C}_1} \wedge 
    \underbrace{(a_1 \vee b_1 \vee b_2)}_{\mathcal{C}_2} \wedge 
    \underbrace{(x_1 \vee x_2 \vee x_3)}_{\mathcal{C}_3}
\end{equation}
We compute $DL(\psi)$ using the following clause entropies.
\begin{align*}
    DL_{cl}(\mathcal{C}_1) &= -\frac{3}{5}log_2(\frac{3}{5})-\frac{2}{5}log_2(\frac{2}{5}) \\
    DL_{cl}(\mathcal{C}_2) &= -\frac{1}{5}log_2(\frac{1}{5})-\frac{4}{5}log_2(\frac{4}{5}) \\&\qquad-\frac{2}{6}log_2(\frac{2}{6}) - 
         \frac{4}{6}log_2(\frac{4}{6}) \\ 
    DL_{cl}(\mathcal{C}_3) &= -\frac{3}{7}log_2(\frac{3}{7})-\frac{4}{7}log_2(\frac{4}{7})
\end{align*}
% $\Upsilon(c_1, \{A\})=-\frac{3}{5}log_2(\frac{3}{5})-\frac{2}{5}log_2(\frac{2}{5})$ \\
% $\Upsilon(c_2, \{A,B\})=\frac{1}{5}log_2(\frac{1}{5})-\frac{4}{5}log_2(\frac{4}{5})-\frac{2}{6}log_2(\frac{2}{6})-\frac{4}{6}log_2(\frac{4}{6}))$ \\ 
% $\Upsilon(c_3, \{X\})=-\frac{3}{7}log_2(\frac{3}{7})-\frac{4}{7}log_2(\frac{4}{7})$ \\
Clauses $\mathcal{C}_1$ and $\mathcal{C}_2$ both mention categorical variable $A$, resulting in an edge $e(\mathcal{C}_1,\mathcal{C}_2)$ in the dual graph of $\psi$. This results in
\begin{equation}
    \begin{split}
            DL(\psi) &= \underbrace{DL_{cl}(\mathcal{C}_1) +
                DL_{cl}(\mathcal{C}_2) + 
                DL_{cl}(\mathcal{C}_3)}_{\text{clause entropies}} + \\
            &\qquad \underbrace{DL_{cl}(\mathcal{C}_2)}_{e(\mathcal{C}_1, \mathcal{C}_2)}+\underbrace{DL_{cl}(\mathcal{C}_1)}_{e(\mathcal{C}_2,\mathcal{C}_1)} \approx 6.21
    \end{split}
\end{equation}
\end{example}

%%%%%%%%%%%%%%%%%%%%%%%%%%%%%%%%%%%%%%%%%%%%%%%%%%%%%%%%%%%%%%%%%%%%%%%%

\section{Problem statement}
\label{Sec:problem statement}
% First we motivate and describe the problem setting. Afterwards we define it more formally.
% playlists -- context
The motivating use case of our work is a problem occurring in the workflow of a music streaming provider, \textit{Tunify}. They have a database of annotated music where each song is represented by a fixed set of discrete valued features. These can be objective (\textit{BPM}, \textit{Year}, \textit{Lyricist}, \ldots) or subjective (\textit{Mood}, \textit{Feel}, \ldots). As one of their services, they provide a predefined selection of playlists. These playlists are represented as logical formulas $\psi$ that can be used as queries on a database to obtain the songs currently matching $\psi$.

% need for safety -- problem
Generating playlists can be time consuming if done manually, as the intended concept can be complex to construct. This is the driving factor to improve the automatic generation of these playlists. Probabilistic circuits are an attractable option to learn them from data, due to their tractable properties and their usability within a setting of only positive and unlabelled data (a PU learning setting where the positive examples are songs that belong to the desired playlist). 

However, \textit{tunify} must also be able to assure their customers that all playlists are safe. For instance, to avoid a black metal song appearing in a playlist intended as happy songs for children, or in a playlist for a funeral home. 
Therefore, to enable easier inspection by a music expert, we wish to extract a formula $\psi$ that acts as a discriminative classifier covering the high density regions of the learned probabilistic circuit. The focus on high density regions allows $\psi$ to only cover the most relevant examples.
\paragraph{Given}
 A probabilistic circuit $\mathcal{M}$ as described in Section~\ref{Sec:theory}, a set of examples $\mathcal{E}$, and a nonzero probability threshold $t$.
 
\paragraph{Objective} 
$\mathcal{E}_{\mathit{HDR}} \subseteq \mathcal{E}$ are examples in the high density regions of $\mathcal{M}$, i.e., the examples ${e \in \mathcal{E}}$ for which ${P_\mathcal{M}(e)\geq t}$. 
Using $\mathcal{E}_{\psi}$ to indicate all examples in $\mathcal{E}$ that are covered by $\psi$, the goal is to find formula $\psi$ such that: 
\begin{itemize}
        \item $\mathcal{E}_{\psi}$ is as close to $\mathcal{E}_{HDR}$ as possible,  
        %\item $\psi$ describes the examples in $\mathcal{E}_{HDR}$. Using $\mathcal{E}_{\psi}$ to indicate all examples in $\mathcal{E}$ that are covered by $\psi$,
        \begin{gather}
         \underset{\psi}{\arg\max} \  F_1(\mathcal{E}_\psi, \mathcal{E}_{HDR}) \label{Objective1} \\ \text{ with } F_1(x,y)=\frac{2\times|x\cap y|}{2\times|x\cap y|+|x-y|+|y-x|}\nonumber
        \end{gather}

        \item whilst minimising the aggregated entropy of $\psi$.
        \begin{gather}
            \underset{\psi}{\arg\min} \ DL(\psi)\label{objective2}
        \end{gather}
    \end{itemize}

%%%%%%%%%%%%%%%%%%%%%%%%%%%%%%%%%%%%%%%%%%%%%%%%%%%%%%%%%%%%%%%%%%%%%%%%
\section{Method}
\label{sec:method}

The input probabilistic circuit $\mathcal{M}$ is a generative model that can be transformed into a discriminative one: given an example $e$, we can compute $P_{\mathcal{M}}(e) \geq t$  (with $t$ the given threshold). However, this discriminative model is not easy to inspect for a domain expert. As described in Section~\ref{sec:bg:prob_circuits}, we can extract a logical formula from a probabilistic circuit, but that formula will be too general. It will also cover examples within the low density regions.

To solve the described problem, we propose to first prune the probabilistic circuit $\mathcal{M}$ in a way that only the high density regions remain. Afterwards, we can extract a logical formula $\psi$ from the pruned circuit and convert it to a CNF. This formula then acts as a discriminative classifier that indicates whether a given example belongs to the high density region of the input probabilistic circuit $\mathcal{M}$, and that is easier to inspect.

We propose a new approach called PUTPUT (Probabilistic circuit Understanding Through Pruning Underlying logical Theories), that prunes a probabilistic circuit while considering the $F_1$-score. This approach consists of two steps.  
The first step eliminates sum node edges of the circuit using existing pruning functions. This results in a circuit only covering the high-density regions. 
A second step eliminates input nodes to further decrease the aggregated entropy of the circuit, whilst keeping the $F_1$-score resulting from the first step as a lower bound.

\subsection{Step 1: pruning sum nodes} 

%There are several ways of pruning a probabilistic circuit.
\paragraph{Pruning functions.} \citet{circuits_Pruning_growing} proposes four pruning functions for a probabilistic circuit. The first approach eliminates random edges of sum nodes in the circuit, while the second approach eliminates sum node edges based on their corresponding weight. Both these approach were identified as less performant so we do not consider them.
The third approach is based on generative significance, pruning sum node edges that contribute the least to the output of the circuit. 
This is represented for each node by its top down probability, which is the probability that the node will be visited when unconditionally drawing samples from the circuit. This function takes as argument the number of edges it has to eliminate. 
This pruning function is shown in Figure~\ref{fig:prune}, applied on the probabilistic circuit from Figure~\ref{fig:intro-box}a.
The fourth approach is based on circuit flows, which works similar to the third approach but first adjusts the sum node weights by conditioning on a given dataset. It shows how many samples from the dataset flow through each node. This pruning function takes as argument the number of edges it has to eliminate and the dataset on which to condition to adjust the weights.

\paragraph{Applying pruning functions.}
PUTPUT first identifies the parameters (i.e., the number of edges to eliminate) of the pruning function that lead to the highest $F_1$-score (Equation~\ref{Objective1}). This can be achieved exhaustively or by using a search function such as golden section search~\cite{golden_section}, which we used in the evaluation. 
The result of step $1$ is a pruned probabilistic circuit such that the $F_1$-score is maximised (see Figure~\ref{fig:prune}).

\subsection{Step 2: pruning input nodes}

\begin{figure}
    \centering
    \includegraphics[width=\columnwidth]{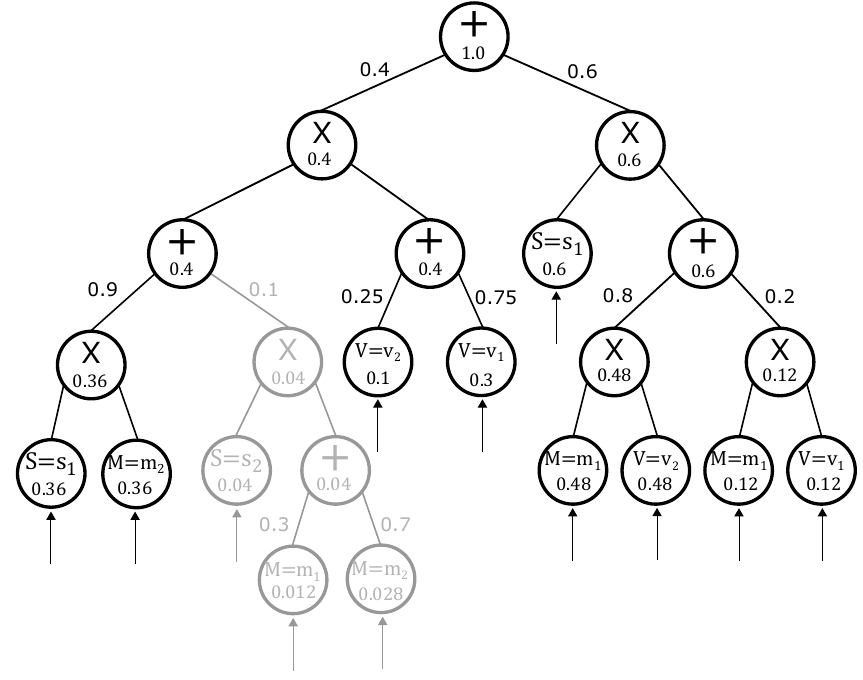}
    \caption{The result after pruning the probabilistic circuit from Figure~\ref{fig:intro-box}a, with the generative significance method set to eliminate the five nodes with the lowest top down probabilities. These values are shown numerically inside the nodes.}
    \label{fig:prune}
\end{figure}

\begin{figure}
    \centering
    \includegraphics[width=\columnwidth]{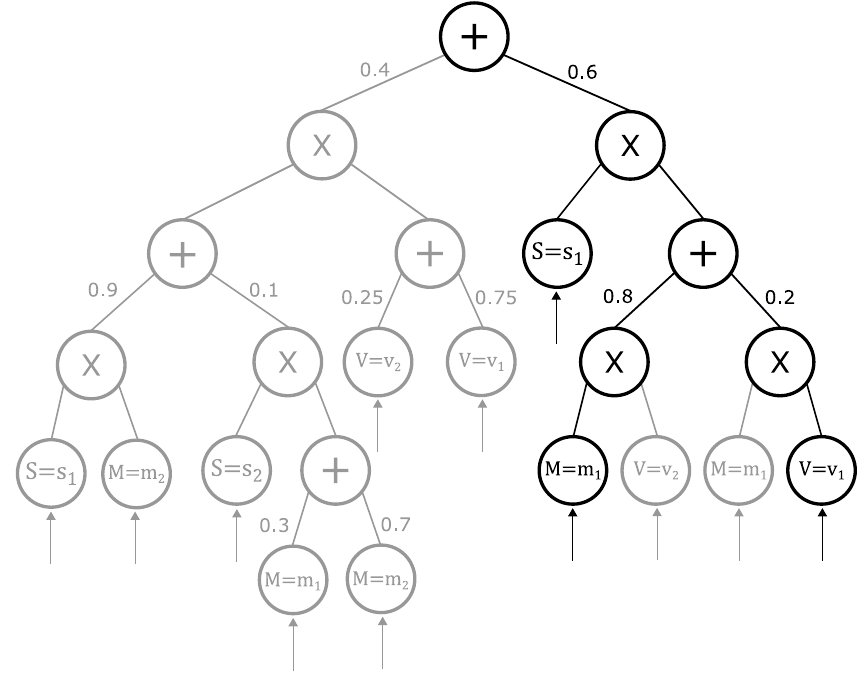}
    \caption{The resulting circuit after applying step 2 of PUTPUT, with $\mathcal{E}_{HDR}=\{(s_1,m_1,v_1), (s_1,m_1,v_2),(s_1,m_2,v_1)\}$}.
    \label{fig:final}
\end{figure}

By correlation with circuit size, the first, previous step of PUTPUT decreases the aggregated entropy. 
The second step decreases this further by considering for each input node whether it is beneficial to be pruned. This differs from the first step which considers only children of sum nodes, while this step also considers input nodes leading into product nodes. The $F_1$-score resulting from the first step is treated as a lower bound while we consider each input node in this second step. Pruning children of a product node may influence whether it is beneficial to prune a node that was previously considered. PUTPUT therefore employs an iterative procedure that reconsiders all nodes until no more changes are made.
The pseudocode for this step is shown in Algorithm~\ref{alg:prune_nodes}. The circuit resulting from step 1 is denoted as $\mathcal{M}'$, while $\mathcal{E}_{\mathcal{M}'}$ are the examples $e\in\mathcal{E}$ for which $P_\mathcal{M}'(e) > 0$, as these are the examples covered by the logical formula derived from $\mathcal{M}'$ (see Section  \ref{sec:bg:prob_circuits}). 
If we apply step 2 on Figure~\ref{fig:prune}, we obtain Figure~\ref{fig:final}.% An example of a pruned circuit with pruned input nodes is shown in Figure~\ref{fig:final}.
\begin{algorithm}
\caption{Step 2: Pruning input nodes in PC}
\label{alg:prune_nodes}
\begin{algorithmic}[1]
\REQUIRE Pruned PC $\mathcal{M}'$, high-density examples $\mathcal{E}_{HDR}$, examples $\mathcal{E}$
\ENSURE $\mathcal{M}'$ with pruned input nodes
\STATE $lower\_bound=F_1(\mathcal{E}_{\mathcal{M}'},\mathcal{E}_{HDR})$   

\REPEAT{
    \FOR{each input node $n$ in $\mathcal{M}'$}
        \STATE $\mathcal{M}''\leftarrow\mathcal{M'} / n$ $\qquad \triangleleft$ Prunes node n
        \IF{$F_1(\mathcal{E}_{\mathcal{M}''},\mathcal{E}_{HDR})\geq lower\_bound$}
            \STATE $\mathcal{M}'\leftarrow \mathcal{M}''$
        \ENDIF
    \ENDFOR}\UNTIL{$\mathcal{M}'$ has not changed}

\RETURN $\mathcal{M'}$
\end{algorithmic}
\end{algorithm}

\section{Evaluation}
\label{sec: evaluation}
The proposed PUTPUT method is empirically evaluated to answer the following research questions: \begin{enumerate}
    \item Which pruning function results in the highest $F_1$-score (step 1)?
    \item Does pruning of the input nodes improve the aggregated entropy (step 2)?
    \item How does PUTPUT perform more generally, including on the music playlist generation task?
\end{enumerate}
% (1) Which pruning function results in the highest $F_1$-score (step 1)? 
% (2) Does pruning of the input nodes improve the aggregated entropy (step 2)?
% (3) How does PUTPUT perform more generally, including on the music playlist generation task?

\subsection{Setup}

Given a dataset of examples $\mathcal{E}$, some labelled positive and some unlabelled, we first learn a probabilistic circuit $\mathcal{M}$ using the Hidden Chow Liu Tree~\cite{hclt} method available in the JUICE package~\cite{juice}. Next, we find the probability threshold $t$ that identifies the high-density regions. As this depends on the dataset, we use the elbow method to determine an appropriate threshold value $t$ (cf. Appendix A). Using this threshold, we determine the high density region examples $\mathcal{E}_{HDR} \subseteq \mathcal{E}$.
Afterwards, we can apply PUTPUT to prune $\mathcal{M}$, eventually leading to the discriminative formula $\psi$.

\subsection{Data}
PUTPUT is evaluated on the following datasets:
\begin{itemize}
    \item \textit{Tunify} provided real world data consisting of $360~000$ songs, annotated with 14 categorical features having 7 to 120 possible values, and a set of intended playlist concepts. From this private data $20$ datasets where constructed, each representing a different playlist concept that has to be learned. We consider three type of concepts.
    \begin{itemize}
    \item \textit{Single playlist concept}: 5 known product playlist concepts with their respective songs, e.g., \texttt{Rock}.
    \item \textit{Disjunctive playlist concepts}: 10 combinations of two known concepts that have a disjunctive form, e.g., \texttt{Rock or Easy Lounge}.
    \item \textit{Exclusive or product concepts}: 5 playlist concepts with a mutually exclusive structure, e.g., \texttt{style=metal XOR feel=aggressive}.
    \end{itemize}
    
    \item black and white images of
    \begin{itemize}
        \item MNIST (10 classes, 2500 examples) \cite{MNIST},
        \item fashionMNIST (10 classes, 2500 examples) \cite{fashionMNIST},
        \item EMNIST letters (26 classes, 2600 examples) \cite{EMNIST}.
    \end{itemize}
    \item the \textit{mushrooms} (2 classes, 8124 examples)~\cite{mushroom} and \textit{splice} (3 classes, 3190 examples)~\cite{splice} datasets of the UCI machine learning repository. These are the most similar to the playlist dataset in terms of the number of features. A dataset with fewer features would generally lead to a lower aggregated entropy.
\end{itemize}

\paragraph{Setup.}
A class in a dataset determines the positive labels. As we consider a PU learning setting, most of the examples are unlabelled. More specifically, for each class in each dataset, we create 10 subsets of the data and randomly label only $5\%$ of the positively labelled examples, resulting in $710$ PUTPUT datasets. We evaluate on the full dataset. 
%with a size of $5\%$ of the class size were randomly selected as training data, with the full dataset as test data.
%This is done to mimic the PU-learning setting where only a limited amount of positive data is given, which is the case in the workflow of \textit{Company}. 
The code of PUTPUT and the experiments on the open-source data are found on \textit{link}\footnote{Will be shared on acceptance}.

\begin{figure}[h!]
\centering
\includegraphics[width=.9\linewidth]{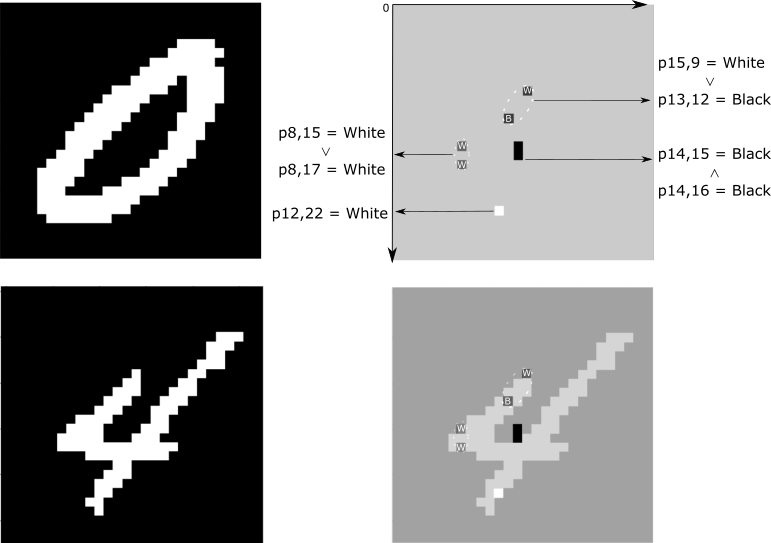} 
\caption{Two MNIST images matching the logical formula $\psi$ extracted by PUTPUT from a learned probabilistic circuit.}
\label{fig: example}
\end{figure}

\paragraph{Example.} 
We provide a small example to motivate our goal of extracting a discriminative classifier that can be understood by a domain expert, and to illustrate its general applicability beyond music playlist generation. We learned a probabilistic circuit on black and white images of MNIST data containing 13 positively labelled examples representing the digit $0$. We then applied PUTPUT to obtain the following logical formula $\psi$.
\begin{equation}
    \begin{split}
        &p_{12,22} = \mathit{White} \wedge p_{14,15} = \mathit{Black} \wedge p_{14,16}=\mathit{Black} \wedge \\
        &(p_{8,15}=\mathit{White} \vee\  p_{8,17}=\mathit{White}) \wedge \\
         & (p_{15,9}=\mathit{White}\vee  p_{13,12}=\mathit{Black})
    \end{split}
\end{equation}
Apparently, the high density region of the learned probabilistic circuit only considers seven of the 784 pixels to predict whether an MNIST digit depicts a $0$. Figure~\ref{fig: example} shows two MNIST digits that match $\psi$ and are part of the high density regions of the probabilistic circuit. The latter can be verified by evaluating the probabilistic circuit for the given image.
In addition to being able to verify and ensure safety, the domain expert can also use description $\psi$ as a starting point to further refine their intended concept.

\subsection{Experiments}

\begin{table*}[t!]
 \caption{Results of experiment 1 after the first step of PUTPUT, showing the average F$_1$-score and circuit size (and standard deviations) in relation to $\mathcal{E}_{HDR}$.}
    \centering
    \begin{tabular}{r|r c l|r c l}
       \multicolumn{1}{c}{ \textbf{Pruning function}} &  \multicolumn{3}{c}{\textbf{F$_1$-score}} &   \multicolumn{3}{c}{\textbf{Circuit size} (\# nodes)} \\
        no pruning & & \ &  & $3470$&$\pm$&$780$ \\ 
         generative significance&$0.306$&$\pm$&$0.24$&$2549$&$\pm$&$692$\\ 
circuit flows&$0.341$&$\pm$&$0.23$&$2058$&$\pm$&$606$\\ 
    \end{tabular}
   
    \label{tab:pruning methods}
\end{table*}

\paragraph{Experiment 1: comparing pruning methods.}
To address research question 1, we evaluate the first step of PUTPUT using the two previously described pruning functions~\cite{circuits_Pruning_growing} on the open-source benchmarks: based on generative significance and based on circuit flows. %The f1-score used in this evaluation is in relation to the target examples.
Table~\ref{tab:pruning methods} shows that pruning by circuit flows is the best approach to optimize the F$_1$-score. Additionally, the resulting circuit is also smaller when using circuit flows. This likely leads to a decrease in aggregated entropy.
In the following experiments, pruning by circuit flows is used as pruning method in the first step of PUTPUT.

\begin{table*}[t!]
\caption{Results of experiment 2, showing average and standard deviation for various metrics, in relation to examples $\mathcal{E}_{HDR}$.}
    \centering
    \begin{tabular}{c|r c l|r c l|r c l|r c l|r c l}
       \multicolumn{1}{c}{}  &  \multicolumn{3}{c}{\textbf{F$_1$-score}} & \multicolumn{3}{c}{\textbf{Aggregated entropy}}&  \multicolumn{3}{c}{\textbf{Precision}}& \multicolumn{3}{c}{\textbf{Recall}} & \multicolumn{3}{c}{\shortstack{\textbf{Circuit size} \\ (\# nodes)}}   \\
        After PUTPUT step 1&$0.341$&$\pm$&$0.23$&$11801$&$\pm$&$49500$&$0.314$&$\pm$&$0.23$&$0.404$&$\pm$&$0.26$&$2058$&$\pm$&$606$\\ 
After PUTPUT step 2&$0.394$&$\pm$&$0.23$&$299$&$\pm$&$459$&$0.314$&$\pm$&$0.23$&$0.419$&$\pm$&$0.25$&$593$&$\pm$&$263$\\  
    \end{tabular}
    
    \label{tab:pruning input}
\end{table*}

\paragraph{Experiment 2: effect of PUTPUT's step 2.}
In the second step of PUTPUT, the input nodes of the probabilistic circuit are pruned to further lower the aggregated entropy.
Table~\ref{tab:pruning input} shows the results of PUTPUT applied on the open-source benchmarks. We conclude that the second step is very significant in decreasing the aggregated entropy. Furthermore, it also increased the F$_1$-score and recall.

\paragraph{Experiment 3: evaluation of PUTPUT.}
In this experiment, we evaluate PUTPUT on all datasets based on the ground truth labels rather than on $\mathcal{E}_{HDR}$. 
We compare against a state-of-the-art concept learning algorithm presented by \citet{kshitij} and that was designed for a similar use case. We call their method \textit{PU+DT} as it learns decision trees in a PU learning context. They propose two approaches to identify examples that are reliably negative before training a decision tree. The first approach uses Rocchio classification~\cite{PU} while the second uses the likelihood. Orthogonally, they propose two approaches to convert the resulting decision tree into a logical formula, denoted as dt-queries or item-queries. In total, this results in four approaches that we include in our comparison.

When transforming the learned decision tree to a logical formula $\psi$, the dt-queries approach of \textit{PU+DT} generates a DNF formula, while item-queries generates a disjunction of multiple CNFs. We transform both of these into a CNF using the well known approach of exploiting distributivity and De Morgan's rule. This may, however, result in an exponentially large CNF for which the aggregated entropy is then also very large. 
Indeed, in our experiments we observed that constructing the CNF for item-queries and the Rocchio+dt approach is not feasible in a reasonable amount of time. This was possible for the likelihood+dt approach, but resulted in an average aggregated entropy of $DL = 2.67\times 10^{8}$ compared to PUTPUT's $DL=188.7$. This is somewhat expected as PUTPUT is specifically designed to minimise this metric. However, PUTPUT also outperforms the other approaches on the primary objective, the F$_1$-score. This confirms empirically its capacity to effectively produce a discriminative classifier that is inspectable by a domain expert. %Furthermore, PUTPUT outperforms the four parameter combinations of \textit{PU+DT} in terms of the primary objective, the F$_1$-score. 
Tabular results per dataset are available in Appendix B.

\begin{figure}
    \centering
    \includegraphics[width=.6\columnwidth]{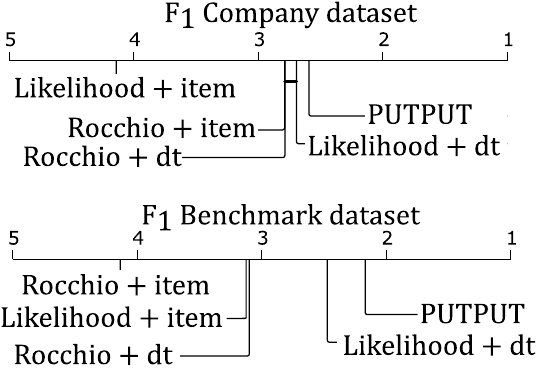}
    \caption{The critical decision diagrams of experiment 3. Right is better. PUTPUT outperforms the \textit{PU+DT} methods on all benchmark datasets.}
    \label{fig:CDs}
\end{figure}

%%%%%%%%%%%%%%%%%%%%%%%%%%%%%%%%%%%%%%%%%%%%%%%%%%%%%%%%%%%%%%%%%%%%%%%%
\section{Related work}
\label{sec:related work}

\paragraph{Explainability in AI} can be tackled in different ways. (1) By limiting to interpretable models, possibly sacrificing accuracy~\cite{xai_expl_tradeoff}; (2) By transforming a learned model into a more interpretable one. For example, by transforming a tree ensemble into a single tree~\cite{xai_explainable_models_compression}, or by compiling a Bayesian network classifier into a logical classifier~\cite{xai_explanations_symbolic}; (3) By transforming part of a model into an interpretable version. This is the approach followed by LIME~\cite{xai_Lime} and SHAP~\cite{xai_SHAP} where linear models are generated around a point of interest; (4) By querying the model to identify edge cases or adversarial examples~\cite{xai_veritas}.
%Such an approach allows a full investigation of the original model, but requires the user to formulate the relevant query~\cite{xai_veritas}. 
In this work we focused on the second strategy, which is useful for domain experts who must inspect and approve entire models before deployment.

\paragraph{Probabilistic circuits.}
\citet{circuits_Pruning_growing} proposed several pruning functions for a probabilistic circuit, functions that we utilise within PUTPUT. They used these functions while devising a prune+grow approach to learn more meaningful probabilistic circuits: insignificant parts are pruned before again growing the remaining part.
%This increased the capacity of the PC that is meaningfully used when generating probabilities \cite{circuits_Pruning_growing}. 
An earlier application of probabilistic circuits in the field of explainable AI is by \citet{xai_probabilistic_sufficient_explanations}, who used them to find explanations that have a high probability of being correct. 
Metrics to measure the uncertainty within probabilistic circuits, to detect out of distribution data~\cite{circuits_know_dont_know}, or to measure fairness for avoiding discrimination~\cite{circuits_fairness}, are two other relevant sources of information while learning a probabilistic circuit. % that provide the user with more information about the concepts the circuit has learned. 

\paragraph{Pattern mining.}
Finding a description of a given set of examples is a problem setting that occurs in the field of data mining. 
KRIMP~\cite{mining_KRIMP} is a pattern mining algorithm that uses the minimum description length (MDL) principle to find a code table that compresses the data. 
Unfortunately, converting this code table into a comprehensive logical formula is not trivial. 
Another approach based on itemset mining is Mistle~\cite{mistle}, which learns and compresses a logical formula based on both positive and negative data, but is not applicable to our PU-learning~\cite{PU} setting. 
A last example of the use of pattern mining to find descriptions, is constraint-based querying to explore Bayesian networks~\cite{mining_bayesian}. The patterns mined in this work are used to answer explorative queries, which are used to explain the Bayesian network representation. 
%In this work pattern mining is used to answer explorative queries, which are used to explain what the Bayesian network is representing. 

\paragraph{Music playlist generation.}
There exist several methods to automatically generate music playlists. However, their learning process does not typically result in an easy to inspect and understand model~\cite{AucouturierP02, ChoiFS16, IreneBZBS19, NabizadehJTY16, TomasiCKCRD23}. Other methods aim to recommend single songs instead of a full playlist, based on preferences, social features or information gathered from wearables~\cite{music_emotion,music_recommendation_music_user_grouping, music_classification, music_came}.

% Automating the generation of playlists, which is the motivating use case of this work, is a recurrent research topic. Recent methods use information registered by wearable physiological sensors to measure the mood of the person~\cite{music_emotion}, use acoustic features of music extracted by digital signal processing~\cite{music_classification} or by examining the MIDI files~\cite{music_recommendation_music_user_grouping}, or use information of the order in which songs are played by the user~\cite{music_came}. \todo{revisit}

%%%%%%%%%%%%%%%%%%%%%%%%%%%%%%%%%%%%%%%%%%%%%%%%%%%%%%%%%%%%%%%%%%%%%%%%

\section{Conclusions}
Motivated by a real world music playlist generation task, we presented an approach to extract a discriminative model (a logic formula) from a probabilistic circuit in a way that is easy to inspect and understand by a domain expert. 
This approach called for 1) a new pruning method (PUTPUT), that prunes low-density regions from the probabilistic circuit; and 2) a new description length for the extracted CNF formula (aggregated entropy). Our experiments have demonstrated the effectiveness of our proposed approach, outperforming competitor methods in terms of the aggregated entropy and F$_1$-score on both the playlist generation use case and open source datasets.

% %%%%%%%%%%%%%%%%%%%%%%%%%%%%%%%%%%%%%%%%%%%%%%%%%%%%%%%%%%%%%%%%%%%%%%%%
% %%% Use this environment to include acknowledgements (optional).
% %%% This will be omitted in doubleblind mode.

% \begin{ack}
% By using the \texttt{ack} environment to insert your (optional) 
% acknowledgements, you can ensure that the text is suppressed whenever 
% you use the \texttt{doubleblind} option. In the final version, 
% acknowledgements may be included on the extra page intended for references.
% \end{ack}

% %%%%%%%%%%%%%%%%%%%%%%%%%%%%%%%%%%%%%%%%%%%%%%%%%%%%%%%%%%%%%%%%%%%%%%%%

%%% Use this command to include your bibliography file.
\clearpage
\bibliography{main}

\clearpage
\section*{APPENDIX}
\subsection*{\textbf{A}. Elbow method}
\label{app:elbow}
This is our implementation of the elbow method. Given set of examples $\mathcal{E}$ and probabilistic circuit $\mathcal{M}$.
Let $\mathcal{L}=[e_1,...,e_n]$ be the examples in $\mathcal{E}$ ordered on the probability generated by PC $\mathcal{M}$, such that $e_1=\argmax_{e\in\mathcal{E}}p_\mathcal{M}(e)$ and $e_n=\argmin_{e\in\mathcal{E}}p_\mathcal{M}(e)$. The elbow method then finds threshold $t$ such that $t=p_\mathcal{M}(e_i)$ with $i=\argmax_{x=1..|\mathcal{E}|}\frac{e_{x+1}-e_x}{e_x-e_{x-1}}<0.3$

\subsection*{\textbf{B}.Tabular results of experiment 3}
\label{app: results}

\begin{table}[h!]
    \centering
    \caption{Tabular results of experiment 3. Aggregated entropy is shown where it was feasible to compute. \# queries represents the number of CNFs in the case of item-queries and PUTPUT, and DNFs in the case of dt-queries.}
    \begin{tabular}{c|r c l|c r c l|r c l|r c l|r c l}
        \multicolumn{1}{c}{} & \multicolumn{3}{c}{\textbf{F1}}&  \multicolumn{4}{c}{\textbf{Aggregated entropy}}  & \multicolumn{3}{c}{\textbf{Precision}}& \multicolumn{3}{c}{\textbf{Recall}}&  \multicolumn{3}{c}{\textbf{\# queries}}\\
        \textit{Single dataset}&&&&&&&&&&&&&&& \\
        
        Rocchio + dt-query&$0.704$&$\pm$&$0.21$&&&-&&$0.606$&$\pm$&$0.22$&$0.878$&$\pm$&$0.19$&$3$&$\pm$&$2$\\ 
Likelihood + dt-query&$0.632$&$\pm$&$0.23$&&&-&&$0.993$&$\pm$&$0.02$&$0.507$&$\pm$&$0.26$&$10$&$\pm$&$7$\\ 
Rocchio + item-query&$0.702$&$\pm$&$0.23$&&&-&&$0.994$&$\pm$&$0.01$&$0.589$&$\pm$&$0.26$&$3$&$\pm$&$2$\\ 
Likelihood + item-query&$0.545$&$\pm$&$0.28$&&&-&&$0.996$&$\pm$&$0.01$&$0.427$&$\pm$&$0.28$&$10$&$\pm$&$7$\\ 
PUTPUT&$0.726$&$\pm$&$0.19$&&$71$&$\pm$&$86$&$0.894$&$\pm$&$0.11$&$0.671$&$\pm$&$0.25$&&$1$&\\ \hline \textit{Disjunctive dataset}&&&&&&&&&&&&&&& \\
Rocchio + dt-query&$0.71$&$\pm$&$0.15$&&&-&&$0.652$&$\pm$&$0.2$&$0.832$&$\pm$&$0.09$&$10$&$\pm$&$8$\\ 
Likelihood + dt-query&$0.763$&$\pm$&$0.15$&&&-&&$0.978$&$\pm$&$0.02$&$0.648$&$\pm$&$0.19$&$17$&$\pm$&$8$\\ 
Rocchio + item-query&$0.767$&$\pm$&$0.11$&&&-&&$0.994$&$\pm$&$0.01$&$0.638$&$\pm$&$0.15$&$10$&$\pm$&$8$\\ 
Likelihood + item-query&$0.666$&$\pm$&$0.21$&&&-&&$0.993$&$\pm$&$0.01$&$0.536$&$\pm$&$0.22$&$17$&$\pm$&$8$\\ 
PUTPUT &$0.732$&$\pm$&$0.17$&&$380$&$\pm$&$462$&$0.816$&$\pm$&$0.2$&$0.693$&$\pm$&$0.19$&&$1$&\\ \hline
\textit{XOR dataset}&&&&&&&&&&&&&&& \\
Rocchio + dt-query&$0.595$&$\pm$&$0.2$&&&-&&$0.467$&$\pm$&$0.21$&$0.953$&$\pm$&$0.05$&$2$&$\pm$&$1$\\ 
Likelihood + dt-query&$0.59$&$\pm$&$0.1$&&&-&&$0.935$&$\pm$&$0.1$&$0.439$&$\pm$&$0.11$&$11$&$\pm$&$5$\\ 
Rocchio + item-query&$0.483$&$\pm$&$0.14$&&&-&&$0.638$&$\pm$&$0.34$&$0.518$&$\pm$&$0.13$&$2$&$\pm$&$1$\\ 
Likelihood + item-query&$0.469$&$\pm$&$0.1$&&&-&&$0.994$&$\pm$&$0.02$&$0.313$&$\pm$&$0.09$&$11$&$\pm$&$5$\\ 
PUTPUT&$0.656$&$\pm$&$0.17$&&$99$&$\pm$&$238$&$0.75$&$\pm$&$0.26$&$0.655$&$\pm$&$0.19$&&$1$&\\ 
\hline \textit{Mnist}&&&&&&&&&&&&&&& \\ 
Rocchio + dt-query&$0.39$&$\pm$&$0.15$&&&-&&$0.771$&$\pm$&$0.17$&$0.277$&$\pm$&$0.15$&$8$&$\pm$&$3$\\ 
Likelihood + dt-query&$0.158$&$\pm$&$0.06$&&&$1.78\times10^8$&&$0.975$&$\pm$&$0.05$&$0.087$&$\pm$&$0.04$&$9$&$\pm$&$2$\\ 
Rocchio + item-query&$0.126$&$\pm$&$0.08$&&&-&&$1.0$&$\pm$&$0.0$&$0.07$&$\pm$&$0.06$&$8$&$\pm$&$3$\\ 
Likelihood + item-query&$0.101$&$\pm$&$0.01$&&&-&&$1.0$&$\pm$&$0.0$&$0.053$&$\pm$&$0.01$&$9$&$\pm$&$2$\\ 
PUTPUT&$0.489$&$\pm$&$0.22$&&$109$&$\pm$&$110$&$0.483$&$\pm$&$0.25$&$0.509$&$\pm$&$0.21$&&$1$&\\ 
\hline \textit{Emnist}&&&&&&&&&&&&&&& \\ 
Rocchio + dt-query&$0.147$&$\pm$&$0.06$&&&-&&$0.435$&$\pm$&$0.21$&$0.095$&$\pm$&$0.05$&$4$&$\pm$&$2$\\ 
Likelihood + dt-query&$0.096$&$\pm$&$0.0$&&&$7.24\times10^6$&&$0.998$&$\pm$&$0.03$&$0.051$&$\pm$&$0.0$&$4$&$\pm$&$1$\\ 
Rocchio + item-query&$0.096$&$\pm$&$0.0$&&&-&&$0.999$&$\pm$&$0.02$&$0.05$&$\pm$&$0.0$&$4$&$\pm$&$1$\\ 
Likelihood + item-query&$0.095$&$\pm$&$0.0$&&&-&&$0.999$&$\pm$&$0.02$&$0.05$&$\pm$&$0.0$&$4$&$\pm$&$1$\\ 
PUTPUT&$0.269$&$\pm$&$0.14$&&$341$&$\pm$&$471$&$0.285$&$\pm$&$0.17$&$0.265$&$\pm$&$0.14$&&$1$&\\ 
\hline \textit{Fashion}&&&&&&&&&&&&&&& \\ 
Rocchio + dt-query&$0.471$&$\pm$&$0.18$&&&-&&$0.514$&$\pm$&$0.25$&$0.503$&$\pm$&$0.15$&$5$&$\pm$&$2$\\ 
Likelihood + dt-query&$0.175$&$\pm$&$0.07$&&&$1.06\times10^8$&&$0.834$&$\pm$&$0.18$&$0.102$&$\pm$&$0.05$&$8$&$\pm$&$2$\\ 
Rocchio + item-query&$0.21$&$\pm$&$0.11$&&&-&&$0.896$&$\pm$&$0.14$&$0.125$&$\pm$&$0.08$&$5$&$\pm$&$2$\\ 
Likelihood + item-query&$0.106$&$\pm$&$0.02$&&&-&&$0.995$&$\pm$&$0.02$&$0.056$&$\pm$&$0.01$&$8$&$\pm$&$2$\\ 
PUTPUT&$0.433$&$\pm$&$0.17$&&$502$&$\pm$&$608$&$0.383$&$\pm$&$0.16$&$0.508$&$\pm$&$0.19$&&$1$&\\ 
\hline \textit{Splice}&&&&&&&&&&&&&&& \\ 
Rocchio + dt-query&$0.177$&$\pm$&$0.04$&&&-&&$0.995$&$\pm$&$0.01$&$0.097$&$\pm$&$0.03$&$26$&$\pm$&$11$\\ 
Likelihood + dt-query&$0.775$&$\pm$&$0.09$&&&$2.89\times10^6$&&$0.707$&$\pm$&$0.15$&$0.896$&$\pm$&$0.1$&$3$&$\pm$&$2$\\ 
Rocchio + item-query&$0.113$&$\pm$&$0.01$&&&-&&$1.0$&$\pm$&$0.0$&$0.06$&$\pm$&$0.01$&$25$&$\pm$&$11$\\ 
Likelihood + item-query&$0.732$&$\pm$&$0.12$&&&-&&$0.728$&$\pm$&$0.17$&$0.811$&$\pm$&$0.2$&$3$&$\pm$&$2$\\ 
PUTPUT&$0.711$&$\pm$&$0.13$&&$8$&$\pm$&$13$&$0.628$&$\pm$&$0.13$&$0.838$&$\pm$&$0.17$&&$1$&\\ 
\hline \textit{Mushrooms}&&&&&&&&&&&&&&& \\ 
Rocchio + dt-query&$0.761$&$\pm$&$0.11$&&&-&&$0.761$&$\pm$&$0.15$&$0.774$&$\pm$&$0.1$&$50$&$\pm$&$20$\\ 
Likelihood + dt-query&$0.986$&$\pm$&$0.01$&&&$2.29\times10^8$&&$0.999$&$\pm$&$0.0$&$0.973$&$\pm$&$0.01$&$9$&$\pm$&$2$\\ 
Rocchio + item-query&$0.819$&$\pm$&$0.08$&&&-&&$1.0$&$\pm$&$0.0$&$0.702$&$\pm$&$0.12$&$50$&$\pm$&$20$\\ 
Likelihood + item-query&$0.97$&$\pm$&$0.02$&&&-&&$1.0$&$\pm$&$0.0$&$0.942$&$\pm$&$0.03$&$9$&$\pm$&$2$\\ 
PUTPUT&$0.89$&$\pm$&$0.1$&&$103$&$\pm$&$133$&$0.918$&$\pm$&$0.09$&$0.892$&$\pm$&$0.16$&&$1$&\\ 
    \end{tabular}
    \label{tab:kshitij}
\end{table}

\end{document}